\documentclass[nohyperref]{article}

\usepackage{microtype}
\usepackage{graphicx}
\usepackage{subfigure}
\usepackage{booktabs} % for professional tables
\usepackage{threeparttable}
\usepackage{placeins}
\usepackage[accepted]{icml2022}

\usepackage[textsize=tiny]{todonotes}

\usepackage{wrapfig}

\usepackage[utf8]{inputenc} % allow utf-8 input
\usepackage[T1]{fontenc}    % use 8-bit T1 fonts
\usepackage{hyperref}       % hyperlinks
\usepackage{url}            % simple URL typesetting
\usepackage{booktabs}       % professional-quality tables
\usepackage{amsmath, amsthm, amsfonts, amssymb, amscd, bm}
\usepackage{multirow}
\usepackage{nicefrac}       %  compact symbols for 1/2, etc.
\usepackage{microtype}      % microtypography
\usepackage{xcolor}         % colors
\usepackage{adjustbox}
\usepackage{algorithm}
\usepackage{algorithmic}
\usepackage{makecell}
\usepackage{xspace}
\usepackage{enumitem}
\newtheorem{proposition}{Proposition}
\usepackage{caption}

\icmltitlerunning{Bayesian Learning with Information Gain Provably Bounds Risk for a Robust Adversarial Defense}

\newcommand{\D}{\mathcal{D}}
\newcommand{\Da}{\mathcal{D}_{\text{adv}}}

\newcommand{\btheta}{\boldsymbol\theta}
\newcommand{\bTheta}{\boldsymbol\Theta}

\newcommand{\hatff}{\hat{\boldsymbol{\phi}}}

\newcommand{\dtoprule}{\specialrule{1pt}{0pt}{\belowrulesep}
            }
\newcommand{\dbottomrule}{%\specialrule{0.3pt}{0pt}{0.4pt}%
            \specialrule{1pt}{0pt}{\belowrulesep}%
            }

\newcommand{\bx}{\textbf{x}}

\newcommand{\bdelta}{\boldsymbol{\delta}}
\def\bxa{\bx_{\text{adv}}}

\def\II{\mathbb{I}}
\def\ie{\textit{i.e.}\xspace}
\def\eg{\textit{e.g.}\xspace}

\begin{document}

\twocolumn[
\icmltitle{Bayesian Learning with Information Gain Provably Bounds Risk for a Robust Adversarial Defense}

\icmlsetsymbol{equal}{*}

\begin{icmlauthorlist}
\icmlauthor{Bao Gia Doan}{sch}
\icmlauthor{Ehsan Abbasnejad}{sch}
\icmlauthor{Javen Qinfeng Shi}{sch}
\icmlauthor{Damith C. Ranasinghe}{sch}
\end{icmlauthorlist}

\icmlaffiliation{sch}{School of Computer Science, University of Adelaide, SA, Australia}

\icmlcorrespondingauthor{Bao Gia Doan}{giabao.doan@adelaide.edu.au}

\icmlkeywords{Machine Learning, ICML}

\vskip 0.3in
]

\printAffiliationsAndNotice{}  % leave blank if no need to mention equal contribution

\begin{abstract}
 We present a new algorithm to learn a deep neural network model robust against adversarial attacks. Previous algorithms demonstrate an adversarially trained Bayesian Neural Network (BNN) provides improved robustness. We recognize the adversarial learning approach for approximating the multi-modal posterior distribution of a  Bayesian model can lead to mode collapse; consequently, the model's achievements in robustness and performance are sub-optimal. Instead, we first propose preventing mode collapse to better approximate the multi-modal posterior distribution. Second, based on the intuition that a robust model should \textit{ignore  perturbations} and only consider the informative content of the input, we conceptualize and formulate an \textit{information gain objective} to measure and force the information learned from both benign and adversarial training instances to be similar. Importantly. we prove and demonstrate that minimizing the information gain objective allows the adversarial risk to approach the conventional empirical risk. We believe our efforts provide a step toward a basis for \textit{a principled method of adversarially training BNNs}. 
Our model demonstrate significantly improved robustness--up to 20\%--compared with adversarial training~\cite{pgd} and Adv-BNN~\cite{advbnn} under PGD attacks with 0.035 distortion on both CIFAR-10 and STL-10 datasets.

\end{abstract}

\section{Introduction}
Deep neural networks (DNNs) have demonstrated \textit{impressive} performance on challenging tasks, such as image recognition~\citep{he2016deep} and natural language processing~\citep{vaswani2017attention}. Despite the impressive performance, DNNs are poor at quantifying the predictive uncertainty and tend to produce overconfident predictions. Consequently, DNNs are shown to be vulnerable to easily crafted perturbations added to the inputs---so-called adversarial examples (AEs)~\citep{Szegedy2014}---to significantly hinder their performance. In image classification tasks, these perturbations are \textit{imperceptible} to human eyes~\citep{fgsm} but can drastically degrade a DNN's performance. There are various methods to find such perturbations in whitebox~\citep{pgd,fgsm,cw,papernot2016transferability, Yuan2020} and blackbox settings~\cite{brendel2017decision, cheng2019sign, chen2020hopskipjumpattack, Vo2022, vo2022query}. Alarmingly, these threats are also shown to be effective in the physical world~\citep{kurakin2018adversarial, eykholt2018robust} and effective in transferring across models to perform \textit{black-box} attacks~\citep{papernot2016transferability, papernot2017practical}. Adversarial perturbations pose a realistic threat for DNN applications and motivate the need to develop robust DNNs.

\textbf{Adversarial Training.~}Despite the immense effort to overcome threats posed by adversarial examples, training a DNN robust against these attacks is challenging.  \citet{athalye2018obfuscated} have shown that one of the most robust defenses against the threat is Adversarial Training~\citep{pgd}. Now, a network is trained with adversarial examples to build robustness against input perturbations post model deployment. But, as mentioned by~\citet{BAL}, the adversarial training algorithm relies on the ``\textit{point estimate}" approach of a deep neural network---a fixed set of network parameters maps the input to the output. 
Essentially, a point estimate with a choice of parameters only defines a single decision boundary that could be easily manipulated with a stronger adversarial input beyond the pre-defined adversarial constraints, \eg maximum norm of perturbations. Alternatively, we can use multiple decision boundaries from a distribution of model parameters and integrate out the effects of parameter choice in the model. That is the premise of Bayesian Deep Neural Network (BNN) learning methods~\cite{welling2011bayesian} aiming to learn a distribution over the model parameters. Now, the output \emph{predictive} distribution is obtained by integrating out the model parameters sampled from their distribution.

\textbf{Bayesian Adversarial Training.~}Motivated by the intuition that removing the effects of the parameter choice can lead to more robust models, \citet{advbnn} proposed adversarial training of BNNs and demonstrated impressive results. However, training BNNs pose a significant challenge; the exact solution of the posterior distribution (\ie the model parameter distribution after observing the data) is \emph{intractable}.
Efforts devoted to developing a suitable inference approach to approximate the posterior involve either using Markov Chain Monte Carlo (MCMC; asymptotically accurate but slow; see \eg \citep{welling2011bayesian} or variational inference (efficient but inaccurate; see \eg \citep{blei2017variational}. For instance, \citet{advbnn} uses a variational method named Bayes by Backprop (BBB)~\citep{blundell2015weight} to approximate the posterior with a unimodal Gaussian distribution. Whilst being an efficient learning algorithm, the challenge faced with such a learning algorithm is the difficulty of capturing the multi-modal aspect of the posterior distribution because the parameters sampled are in the proximity of the mode of the distribution.

\textbf{Our Hypothesis.~}We are motivated to explore the potential robustness gains attainable from an adversarial training algorithm for a Bayesian Deep Neural Network capable of approximating the multi-modality of the posterior. We hypothesize a model (1) learning a better approximation of the parameter distribution that (2)
{gains the same information from the given input and its adversarial counterpart} is more robust.

\textbf{Our Contribution.~} In this paper, to achieve (1), we \textit{combine} adversarial training with an inference approach to faithfully capture the posterior distribution of parameters. The learning of an approximate multi-modal posterior is not new, inspired by~\citet{Liu2016}, we employ Stein Variational Gradient Descent (SVGD) that encourages \emph{diverse} sampling from the posterior. By utilizing the SVGD approach, to achieving (2), we design an \emph{Information Gain} (IG)\footnote{also known as \textit{Mutual Information}~\cite{houlsbyBayesianActiveLearning2011, galDeepBayesianActive2017}} objective. We summarize our contributions below:

\setlist{leftmargin=12pt,itemsep=6pt,topsep=2pt}
\begin{itemize}
 
    \item We propose a novel method to learn a BNN robust against adversarial attacks by utilizing
    SVGD to generate parameter particles that are parallelly trained to be \textit{as diverse as possible} whilst \textit{maintaining the same measure of information content learned from benign and adversarial instances}. Our learning approach enables the model to both reduce the effect of single parameter choice and learn the invariant patterns common between the training dataset and its corresponding adversarial samples.
    
    \item To maintain the same measure of information content learned from both benign and adversarial training instances, we %exploit the diversity in the SVGD inference method to 
    formulate an information gain (mutual information) objective. Our proposed objective reinforces the minimization of the empirical adversarial risk by forcing the information gained, learning from the benign and adversarial samples, to be similar. 
    
    \item We prove, \textit{minimizing the information gain objective} allows the adversarial risk to approach the empirical risk minimization bound. Simply, the risk of misclassification of an adversarial example is now the same as the risk of misclassifying a benign sample. This is the first time such a bound is formally derived; this is significant because it provides a theoretically justified approach to reducing the uncertainty associated with adversarial examples.

    \item Comprehensive evaluations on a set of neural architectures and datasets demonstrate our approach achieves 
    significant improvement in robustness compared to previous methods.
\end{itemize}

\section{Background \& Related Work}

\begin{figure}[h!]
    \centering
    \includegraphics[width=\linewidth]{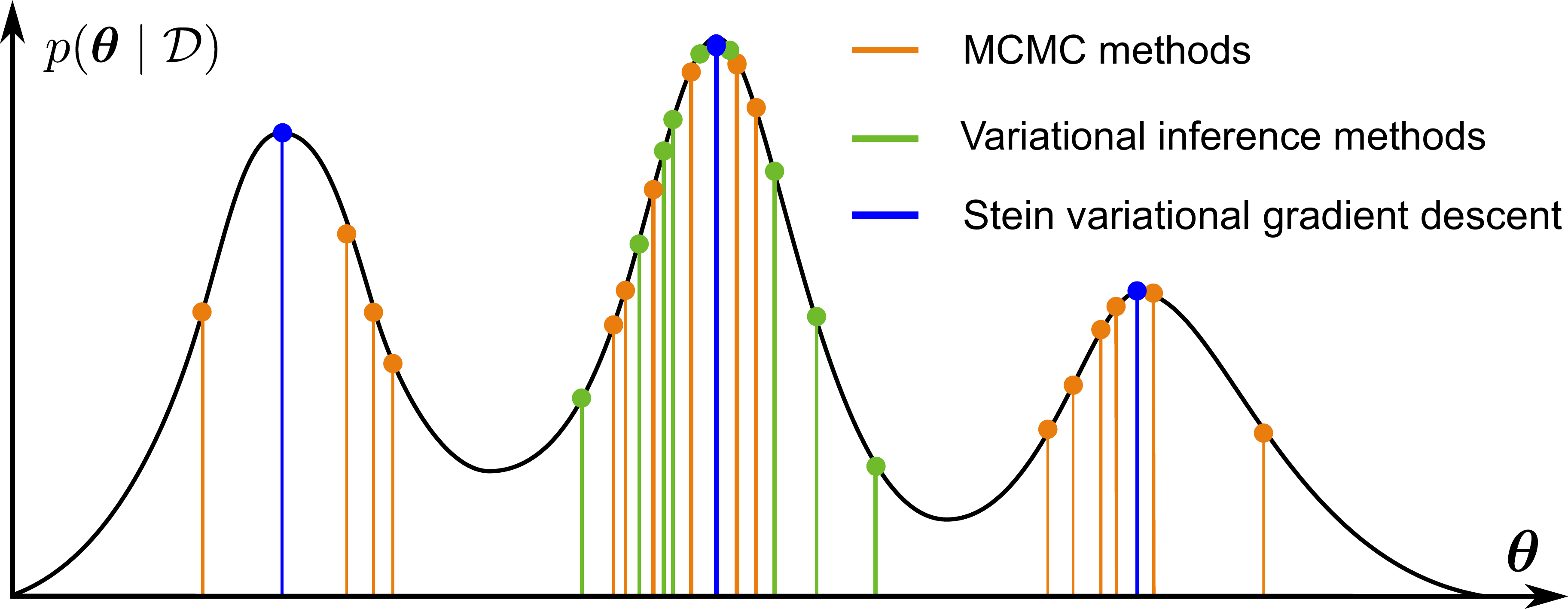}
    \caption{Different techniques to sample the posterior.}
    \label{fig:posterior}
\end{figure}

\noindent\textbf{Primer on Bayesian Learning}. Given a dataset $\mathcal{D} = \left\{\bx_i,y_i\right\}_{i=1}^N$,  a Bayesian Neural Network (BNN) aims to learn  
the \textit{posterior} distribution: $p(\boldsymbol{\theta} \mid {\mathcal{D}}) = \frac{p(\mathcal{D} \mid \boldsymbol{\theta}) p(\boldsymbol{\theta})}{p(\D)}$ 
given the prior distribution $p(\boldsymbol{\theta})$.
However, the exact solution for the posterior is often \textit{intractable} since the deep neural networks are complex distributions and infeasible due to 
the high dimensional integral of the denominator even for moderately sized networks in the context of deep learning~\citep{blei2017variational}. 
In addition, the true Bayesian posterior is usually a complex multimodal distribution \citep{izmailov2021bayesian} as illustrated in Figure~\ref{fig:posterior}. 

Variational inference, which relies on another parametric distribution, is too restrictive to resemble the true posterior and suffers from mode collapse~\citep{izmailov2021bayesian}. 
On the other hand, \citet{wang2019nonlinear, Liu2016} proposed a provable general purpose variational inference algorithm named Stein Variational Gradient Descent (SVGD) that transports a set of parameter particles, encouraged to be diverse, to fit the true posterior distribution; this approach can be beneficial for achieving higher performance and approximating the true posterior distribution.  
The visualization for different techniques to sample the posterior is displayed in Figure~\ref{fig:posterior}.

\noindent\textbf{Adversarial Attacks}. Attackers can add carefully crafted noise (perturbations) to the input image to fool the classifier at the inference stage.  
In general, the goal of the attacker---described in Equation~\eqref{eq:adv}---is to degrade the performance of a neural network by crafting $\bdelta$, such that:
\begin{equation}
    \label{eq:adv}
    \max _{\|\boldsymbol{\delta}\|_{p}<\varepsilon_{\max }} \ell(f(\mathbf{x}+\boldsymbol{\delta} ; \btheta), y)
\end{equation}
where, $p$ is the norm,
$\varepsilon_{\max}$ is the maximum attack budget (perturbation), $\ell$ is the loss function (typically cross-entropy), $f$ is the network, $\mathbf{x}$ is the input, $\btheta$ is the network parameter, and $y$ is the ground-truth label.

For a PGD~\citep{pgd} attack, an attacker starts from $\bx^0=\bx_{o}$ and conducts projected gradient descent iteratively to update the adversarial example following the Equation~\eqref{eq:pgd}:
\begin{equation}
    \label{eq:pgd}
    \bx^{t+1}=\Pi_{\varepsilon_{\max}}\left\{\bx^{t}+\alpha \cdot \operatorname{sign}\left(\nabla_{\bx} \ell\left(f\left(\bx^{t} ; \boldsymbol{\theta}\right), y_{o}\right)\right)\right\}
\end{equation}
where $\Pi_{\varepsilon_{\max}}$ is the projection to the set $\left\{\bx \mid\left\|\bx-\bx_{o}\right\|_{\infty} \leq \varepsilon_{\max}\right\}$

Among all the attack methods, we decided to apply PGD in our experiments because: i)~PGD~\citep{pgd} is regarded as the strongest attack in terms of the $\ell_\infty$ norm and ii) it gives us direct control over the distortion by changing $\varepsilon_{max}$.

\noindent\textbf{Adversarial Defenses}.  Significant research efforts describe methods to mitigate this threat, such as distillation~\citep{jsma}, input denoising~\citep{song2017pixeldefend} or feature denoising~\citep{xie2019feature}, curious readers can find more from~\citep{kurakin2018adversarial}. Among these methods, adversarial training~\citep{pgd} and its variants are shown to be one of the most effective and popular methods to defend against adversarial attacks~\cite{athalye2018obfuscated}. The goal of adversarial training is to incorporate the adversarial search within the training process and, thus, realize robustness against adversarial examples at test time. This is achieved by solving the following optimization problem: 
\begin{equation}
    \label{eq:advtraining}
    \btheta^* = \underset{\btheta}{\arg \min }\underset{(\bx,y)\sim D}{\mathbb{E}}\left\{\max _{\|\bdelta\|_{p}<\varepsilon_{\max }} \ell(f(\mathbf{x}+\bdelta ; \btheta), y)\right\}
\end{equation}
where $\D$ is the training data. An approximate solution for the inner maximization can be realized by generating the PGD adversarial examples from Equation~\eqref{eq:pgd} and then minimizing the classification loss based on the generated PGD adversarial examples. 

Recent works~\citep{atsague2021mutual,zhu2020learning} also incorporate different variants of mutual information into their methods to realize a robust neural network. However, the mutual information is still utilized in a traditional ``point-estimate'' neural network setting; hence, the achieved robustness is marginal compared with traditional adversarial training. In contrast, this paper focuses on formulating mutual information (information gain) in a Bayesian neural network and theoretically prove that Bayesian adversarial learning with information gain allows the adversarial risk to approach the conventional risk.

\noindent\textbf{Prior Art on Bayesian Defenses.~}
Bayesian Neural Networks were proposed to detect adversarial attacks~\citep{feinman2017detecting, smith2018understanding}. Recently, \cite{Carbone21} prove the robustness of BNNs to gradient-based adversarial attacks in the large data and overparameterized limit, while certified adversarial robustness on small $\varepsilon_{\max}$ was shown in~\cite{wicker21}. On the other hand,
\citet{BAL} and \citet{advbnn} tried to combine Bayesian learning with adversarial training. \citet{BAL} present a method to jointly sample from the model's parameter posterior and the distribution of adversarial samples given the current parameter posterior to learn robust BNNs.~\citet{advbnn} further developed the direction proposed in Random Self-Ensemble (RSE)~\citep{RSE} to build an adversarially-trained Bayesian neural network method named Adv-BNN that can scales up to complex data by adding noise to each weight instead of input or hidden features as in RSE~\citep{RSE}. Adv-BNN also incorporates adversarial training to learn a \textit{variational posterior distribution} to further improve model robustness against strong adversarial examples with large $\varepsilon_{\max}$. However, using the variational inference method is likely to lead to mode collapse and limit the performance of the BNN~\citep{izmailov2021bayesian} as we discussed earlier and demonstrate in our experiments in Section~\ref{sec:experiments}. 

In this work, we propose exploring SVGD \citep{Liu2016} as a Bayesian inference method to achieve a better approximation for the multi-modal posterior of a BNN. Using this approach, it is also easy to convert a traditional neural network to a Bayesian counterpart without much effort to modify the traditional neural network architecture. 
Further, by employing the \emph{repulsive force} for encouraging exploration in the parameter space, we conceptualize the Information Gain in Bayesian learning to bound the difference of empirical risk versus the adversarial risk to further improve the robustness on strong adversarial examples.

\section{Method}

Our method combines adversarial training with an inference approach to faithfully capture the posterior distribution of parameters and formulate a new information gain objective in the setting to achieve a provably bounded adversarial risk to, hopefully, achieve a robust adversarial defense. We describe our formulation in what follows.

\subsection{Bayesian Formulation for Adversarial Learning}
\label{sec:bayesian_formulation}
In contrast to a point estimate learned in conventional deep learning models, in Bayesian learning, the posterior of the parameters is obtained using the Bayes rule \ie:
\begin{equation*}
p(\btheta\mid\D)={\prod_{(\bx,y)\sim\D}p(y\mid\bx,\btheta)p(\btheta)}/{Z}
\end{equation*}
where $Z$ is the normalizer. Similarly, for the dataset of adversarials instances $\Da$, we obtain a corresponding  posterior $p(\btheta\mid\Da)$.
We consider $p(y\mid\bxa,\btheta)=\text{softmax}(f(\bxa;\btheta))$ where $f$ is a deep neural network. 
For adversarial dataset $\Da$, since adversarial examples can be generated from their corresponding benign instances, 
we can obtain $\Da$ during adversarial training by applying adversarial attacks such as PGD attacks.
 However, we acknowledge that PGD attacks cannot be directly applied in a BNN setting~\cite{advbnn}. Hence, to account for the uncertainty of BNNs, we utilize Expectation-over-Transformation (EoT)~\citep{athalye2018synthesizing} approach to deploy an EoT PGD attack described in Equation~\eqref{eq:a_pgd}; previously shown in~\citet{rol2019comment}. This attack is more tailored for BNNs due to the fact that it achieves a more representative approximation to estimate the gradient and is formulated as:
{\small\begin{align}\vspace{-3mm}
    \label{eq:a_pgd}
    \bx^{t+1}=\Pi_{\varepsilon_{\max}}\left\{\bx^{t}+\alpha \cdot \operatorname{sign}\left({\mathbb{E}_{\btheta}}\left[\nabla_{\bx} \ell\left(f\left(\bx^{t} ; \btheta\right), y_{o}\right)\right]\right)\right\}.
    % \notag \\    \vspace{-3mm}
\end{align}}
However, the posterior distribution, in general, is intractable and we need to resort to approximations. In particular, we propose utilizing Stein variational gradient descent (SVGD) \citep{Liu2016} which provides an approach to learning multiple \emph{particles} for parameters in parallel to approximate the true posterior. SVGD uses a repulsive factor to encourage the diversity of parameter particles to prevent mode collapse. This diversity enables learning multiple models to represent various patterns in the data. Collectively, the patterns are less vulnerable to adversarial attacks. 
Using $n$ samples from the posterior (\ie parameter particles) the variational bound is minimized when gradient descent is modified as: % \begin{equation}
 \begin{multline}
 \label{eq:grad_update}
     \btheta_i  =   \btheta_i - \epsilon_i \hatff{}^*(\btheta_i)  
\text{\quad with} \\
\hatff{}^*(\btheta) = \sum_{j=1}^n\big[  k(\btheta_j, \btheta)  \nabla_{\btheta_j} \ell(f(\bxa;\btheta_j),y) \\  \,\,\,-\frac{\gamma}{n}\nabla_{\btheta_j} k(\btheta_j, \btheta)\big]\,.
\end{multline}
 Here, $\btheta_i$ is the $i$th particle, $k(\cdot, \cdot)$ is a kernel function that measures the similarity between particles, $\gamma$ a hyper-parameter and
 $\ell(\cdot,\cdot)$ is the cross entropy loss. Notably, the kernel function encourages the particles to be dissimilar to capture more diverse samples from the posterior and $\gamma$ controls the trade-off between the diversity of the samples versus the minimization of the loss.

Further, given the test data point $\bx^*$, we can approximate the posterior using the Monte Carlo samples as
\begin{align*}
    p ( y^* \mid & \mathbf{x}^*, \Da ) = \int p(y^* \mid \mathbf{x}^*, \boldsymbol{\theta}) p(\boldsymbol{\theta} \mid \Da) d \boldsymbol{\theta} \\
    & \approx \frac{1}{n} \sum_{i=1}^{n} p(y^* \mid \mathbf{x}, \boldsymbol{\theta}_{i}), \quad \boldsymbol{\theta}_{i} \sim p(\boldsymbol{\theta} \mid \Da)\,,
\end{align*}
where $\btheta_i$ is an individual parameter particle. 

Importantly,  in the adversarial setting, it is critical to take parameter samples that represent different modes of the distribution that may not have the same vulnerabilities towards perturbations.
The adversarial instances are generally known to exploit the particular patterns learned by the parameters~\citep{papernot2016limitations}. When integrating out the parameters as in the Bayesian setting, especially under the diverse parameter particles in our approach, we implicitly remove the vulnerabilities that could arise from a single choice of a parameter.

\subsection{Conceptualizing Information Gain for Bayesian Learning}

Using the Bayesian setting we employ, we can formulate a notion of information gain that captures the impact of adding a new training instance to a dataset on the distribution of the parameters. The information gain can be defined as (see Appendix~\ref{sec:appd_proofdefinition}):
\begin{equation}
\label{eq:infogain}
\text{IG}(\bx,y;\bTheta)	=
\mathbb{H}[\mathbb{E}_{\btheta}[y|\bx,\D]] - \mathbb{E}_{\btheta}[\mathbb{H}[y|\bx,\D]]
\end{equation}

This formulation quantifies an instance's informativeness for a model given the training set. Intuitively, the information gained from an instance is proportionate to the reduction in the expected entropy of the predictive distribution. 

Our conjecture is that \textit{a robust neural network quantifies the information gain from an observation the same as its adversarial counterpart} \ie $\mathbb{E}_{(\bx,y)\sim \D}[\text{IG}(\bx,y;\bTheta)]=\mathbb{E}_{(\bx_\text{adv},y) \sim \D_\text{adv}}[\text{IG}(\bx_{\text{adv}},y;\bTheta)]$. In other words, a robust model ignores the perturbations and only considers the informative content of the input.
We will employ these concepts in the following learning formulation.

\subsection{Formulate Learning a Robust Network Using Information Gain}

We formulate the objective of our training to: 
\begin{enumerate}[topsep=-1pt,itemsep=0ex,partopsep=0ex,parsep=0ex]
    \item Learn the posterior from the \emph{adversarial} dataset. Since we use SGVD, this corresponds to learning multiple parameter particles. 
    This amounts to minimizing the loss subject to the repulsive constraint, \ie $\mathbb{E}_{(\bx_\text{adv},y) \sim \Da}\left[\mathbb{E_{\btheta\sim p(\btheta\mid\Da)}}[\ell(f(\bxa;\bTheta),y)]\right]$. Since the adversarial dataset  is generated while training the model, it depends on the particle chosen and its parameters. 
    With SGVD, we ensure the samples are diverse, and each parameter particle explores a different pattern in the input. 
    
    \item Achieve comparable information gain from both the given dataset and the adversarials. Thus, ensuring: i)~the information gained from data and adversarial examples is encouraged to be the same, \ie $\mathbb{E}_{(\bx,y)\sim \D}[\text{IG}(\bx,y;\bTheta)]=\mathbb{E}_{(\bx_\text{adv},y) \sim \D_\text{adv}}[\text{IG}(\bx_{\text{adv}},y;\bTheta)]$; ii)~the model to be \textit{not} biased towards learning from the adversarial instances; and iii)~the receptive fields are active for similar and prominent features.
\end{enumerate}

To this end, we formulate the problem as a constrained optimization: 
{\small\begin{align}
\underset{\btheta}{\min} &\quad\mathbb{E}_{(\bx_\text{adv},y) \sim \Da}\left[L(\bxa,y;\bTheta)\right] \label{eq:first_loss}\\
\text{s.t.} & \quad\mathbb{E}_{(\bx,y)\sim \D}[\text{IG}(\bx,y;\bTheta)]=\mathbb{E}_{(\bxa,y) \sim \Da}[\text{IG}(\bxa,y;\bTheta)] \notag
\end{align}}
where $L(\bxa,y;\bTheta)=\mathbb{E_{\btheta\sim p(\btheta\mid\Da)}}[\ell(f(\bxa;\btheta),y)]$.
Combining the above concepts using the Lagrangian method, we have the following objective:
{\small\begin{align}
    \label{eq:main_obj}
    L_{\text{IG}}(\bTheta)\!=\! 
    % \frac{1}{n} \sum_{l=1}^{n} &\ell\left(f\left(\bxa ; \btheta_{l}\right), y\right)
    L(\bxa,y;\bTheta)
    % \\ \nonumber &
    \!+\!\lambda \!\left| \text{IG}(\bx,y;\bTheta) - \text{IG}(\bx_\text{adv},y;\bTheta) \right|
\end{align}}
where we use the Monte Carlo sampling using the particles to estimate the expectations.
Subsequently, this learning objective $L_{\text{IG}}(\bTheta)$ is optimized using the SVGD method in Equation~\eqref{eq:grad_update} mentioned earlier in Section~\ref{sec:bayesian_formulation}. 
Effectively, using this approach, we compute a posterior in a constrained space defined by the information gain criteria. 
Since the space is constrained, the likelihood of findings "particles" in the posterior that are more robust increases.
We summarize our proposed robust Bayesian learning approach in  Algorithm~\ref{alg:alg1}. Here, following~\citet{Liu2016}, we use the RBF kernel $k(\btheta, \btheta')=\exp \left(-{\left\|\btheta-\btheta^{\prime}\right\|^{2}}/{(2 h^{2})}\right)$ and take the bandwidth $h$ to be the median of the pairwise distances of the set of parameter particles at each iteration.

\begin{algorithm}[h!] % 
\caption{Information Gain-BNN (IG-BNN)}  
\label{alg:alg1}
\begin{algorithmic}[1]
\STATE {\bf Input:} A set of initial parameter particles $\{\btheta_i^0\}_{i=1}^n$, observation data $\D$. 
\STATE {\bf Output:} A set of parameter particles $\bTheta:=\{\btheta_i\}_{i=1}^n$ that approximates the true posterior distribution $p(\btheta\mid\Da)$  
\FOR{$(\bx,y)\sim p(\D)$}
\STATE $\bxa\gets\bx$
\FOR{$t=1\to T$} 
\STATE $\begin{aligned}
\bxa=\Pi_{\varepsilon_{\max}}\left\{\right.\bxa&+\alpha \cdot \operatorname{sign}\left(\right. \\
&\mathbb{E}_{\btheta}\left[\nabla_{\bx} \ell\left(f\left(\bxa ; \btheta_j\right), y\right)\right]\left.\right)\left.\right\}
\end{aligned}$

\COMMENT{Generate Adversarial (Eq.~\eqref{eq:a_pgd})}
\ENDFOR
\FOR{$i=1\to n$}
\STATE $\btheta_i  \gets   \btheta_i - \epsilon_i \hatff{}^*(\btheta_i, \btheta_j) $ 
\text{\quad with}
$\hatff{}^*(\btheta_i,\btheta_j) = \sum_{j=1}^n\big[  k(\btheta_j, \btheta_i)  \nabla_{\btheta_j} L_{\text{IG}}(\bTheta) - \frac{\gamma}{n}\nabla_{\btheta_j} k(\btheta_j, \btheta_i)\big]$
\STATE $\epsilon_i$ is the step size at the current iteration, $k(\btheta,  \btheta')$ is a positive definite kernel that specifies the similarity between $\btheta$ and $\btheta'$, $L_\text{IG}$ is the main objective (Eq.~\eqref{eq:main_obj}), $\gamma,\lambda$ is the weight to control the \textit{repulsive force} that enforces the diversity among parameter particles and IG objective respectively, $\ell$ is the cross-entropy loss function.
\ENDFOR
\ENDFOR
\end{algorithmic}
\end{algorithm}

\subsection{A Relation between Adversarial and Observational Training}

A typical machine learning approach minimizes the empirical risk to learn. There are theoretical and empirical studies on the relation between the empirical risk and the true risk that measures the generalization ability of a learning algorithm. Generalization bounds such as Rademacher complexity or VC dimension for classical approaches or more recent studies for deep learning (see \eg \cite{generalisation_deep}) underpin the theoretical framework for machine learning. 

Notably, the relation between the risk when using samples from the \textit{observational distribution} (\ie the given dataset) versus when using their adversarial counterparts remains unexplored. \textit{It is important, because, while adversarial training has been commonly used, the impact of using such an approach on generalization with respect to the true data distribution is unknown}. We consider a Bayesian model with no specific assumption on the distribution of either the adversarial examples or the perturbations to provide a generic defense approach. The only major assumption we make for the following adversarial risk bound is that the distribution of the data and the corresponding adversarial are sufficiently close. That is a mild assumption when we consider the adversarial instances are obtained from small perturbations of the given training dataset. Thus, we are interested in finding the bound of $\left|R_{\text{adv}}-R\right|$ where
\begin{equation*}
R=\mathbb{E}_{\btheta}\left[\mathbb{E}_{(\bx,y)\sim\D}\left[ \mathbb{E}_{y'\sim p(y\mid\bx,\btheta)}\left[\mathbb{I}(y=y')\right]\right]\right]     
\end{equation*}
is the empirical risk, and
\begin{equation*}
R_{\text{adv}}=\mathbb{E}_{\btheta}\left[\mathbb{E}_{(\bxa,y)\sim\Da}\left[ \mathbb{E}_{y'\sim p(y\mid\bxa,\btheta)}\left[\mathbb{I}(y=y')\right]\right]\right]
\end{equation*}
is the risk of the adversarial examples.  Once we can obtain these, we can simply obtain the overall generalization and robustness bound. The following proposition summarizes our findings.

\begin{proposition}\label{prop1}
The risk of a classifier when trained on the observed training set denoted by $R$ versus when trained with adversarials denoted by $R_{\text{adv}}$ is bounded as
\begin{equation*}
\begin{aligned}
\left| R_{\text{adv}}-R \right| &\leq 1 -\mathbb{E}_{(\bx,y)\sim\D} \Bigg[ \exp \bigg(\big( \mathbb{E}_{\btheta}[r_{\btheta}(\bx,\bxa,y)] \\
&- \lambda\left|\mathbb{E}_{\btheta}[\text{IG}(\bx,y;\bTheta)]-\mathbb{E}_{\btheta}[\text{IG}(\bxa,y;\bTheta)]\right|\big) \bigg) \Bigg],
\end{aligned}
\end{equation*}

where $r_{\btheta}(\bx,\bxa,y)=\sum_{c}^{K}p(y=c\mid\bx,\btheta)\log(p(y=c\mid\bxa,\btheta))$, $\lambda\geq 0$ and $\bxa$ denotes the adversarial example obtained from $\bx$.
\end{proposition}

\emph{Sketch of the Proof.} We simplify the difference between the risks by considering that the difference between individual mistakes is smaller than their product, \ie $$
\begin{aligned}
\mathbb{E}_{y_1\sim p(y\mid\bx,\btheta)}\left[\mathbb{E}_{y_2\sim p(y\mid\bxa,\btheta)}\left[\mathbb{I}[y\neq y_1]-\mathbb{I}[y\neq y_2]\right]\right] \\
\leq \mathbb{E}_{y'\sim p(y\mid\bxa,\btheta)}\left[\mathbb{E}_{y'\sim p(y\mid\bxa,\btheta)}\left[\mathbb{I}[y_1\neq y_2]\right]\right]\\ \leq 1-\sum_{c=1}^K p(y=c\mid\bx,\btheta)p(y=c\mid\bxa,\btheta)\,.
\end{aligned}$$ 

We then use Jensen's inequality when using $\exp(\log(\cdot))$ to obtain the upper bound. The complete proof is provided in the Appendix~\ref{sec:appd_proof}. We can see that the difference between the empirical risk and the adversarial risk is minimized when the upper bound is minimized. Hence, to minimize the upper bound, our main learning objectives are to:
\begin{enumerate}
    \item \textit{Minimize cross entropy for the adversarial examples}. This corresponds to matching the prediction from the adversarial data to that of the observations. Since $(\bx,y)$ is given in the training, we simply minimize the entropy of the adversarial examples. This corresponds to using a cross-entropy loss in Eq.~\eqref{eq:first_loss}.
    
    \item \textit{Minimize the difference between the information gained from the dataset and its adversarial counterparts}. In addition to individual predictions, the information gained from each instance (\ie the benign and its adversarial) has to have a similar impact in terms of how it changes the network parameters.
    % ~\damith{What is both here? where are the two?} 
\end{enumerate}
Notably, since we know $1-\exp(-z)\leq z$, to avoid computational instabilities and gradient saturation, we consider minimizing the upper bound without the exponential function.

Our proposed algorithm is summarized in Algorithm~\ref{alg:alg1}.

\section{Experimental Results}
\label{sec:experiments}
In this section, we verify the performance of our proposed method (IG-BNN) with other baselines in the literature on two popular and standard vision tasks. We use the CIFAR-10~\citep{cifar10} dataset---
a popular benchmark used to evaluate the robustness of a DNN in previous works~\citep{pgd, athalye2018obfuscated}. However, it is also known that adversarial training becomes increasingly hard for high-dimensional data~\citep{schmidt18}. Therefore, to further evaluate the robustness of our method, we conduct an experiment on a high dimensional dataset---STL-10~\citep{stl10} with 5,000 training images and 8,000 testing images with images of $96\times96$ pixels.

In all experiments, we utilized the same networks used in the adversarial training BNN method, Adv-BNN~\citep{advbnn} to fairly compare the results. Specifically, we used the VGG-16 network architecture for CIFAR-10 and the smaller ModelA network for STL-10 used in~\citet{advbnn}. The number of PGD steps and the attack budgets used for training and testing are also set to be the same for a fair comparison---see Appendix~\ref{sec:appd_hyper}  Table~\ref{table:hyperparameters}. Notably, we also conduct the experiment with a larger number of PGD steps in the Appendix~\ref{sec:appd_PGDsteps}, and Figure~\ref{fig:PGDsteps} confirm that 20-step is enough for the EoT PGD attack to reach its full strength.

Because our proposed method evaluates the  robustness of a Bayesian learning method based on Adversarial Training, 
the traditional Adversarial Training (Adv. Training)~\citep{pgd} and adversarially trained Bayesian defense,  Adv Bayesian Neural Network (Adv-BNN)~\citep{advbnn} are good baselines for comparisons. In addition, we also compare our method with networks trained with no defenses (No Defense) as well as Bayesian Neural Networks trained for the tasks.

\begin{figure}[ht!]
    \centering
    \includegraphics[width=\linewidth]{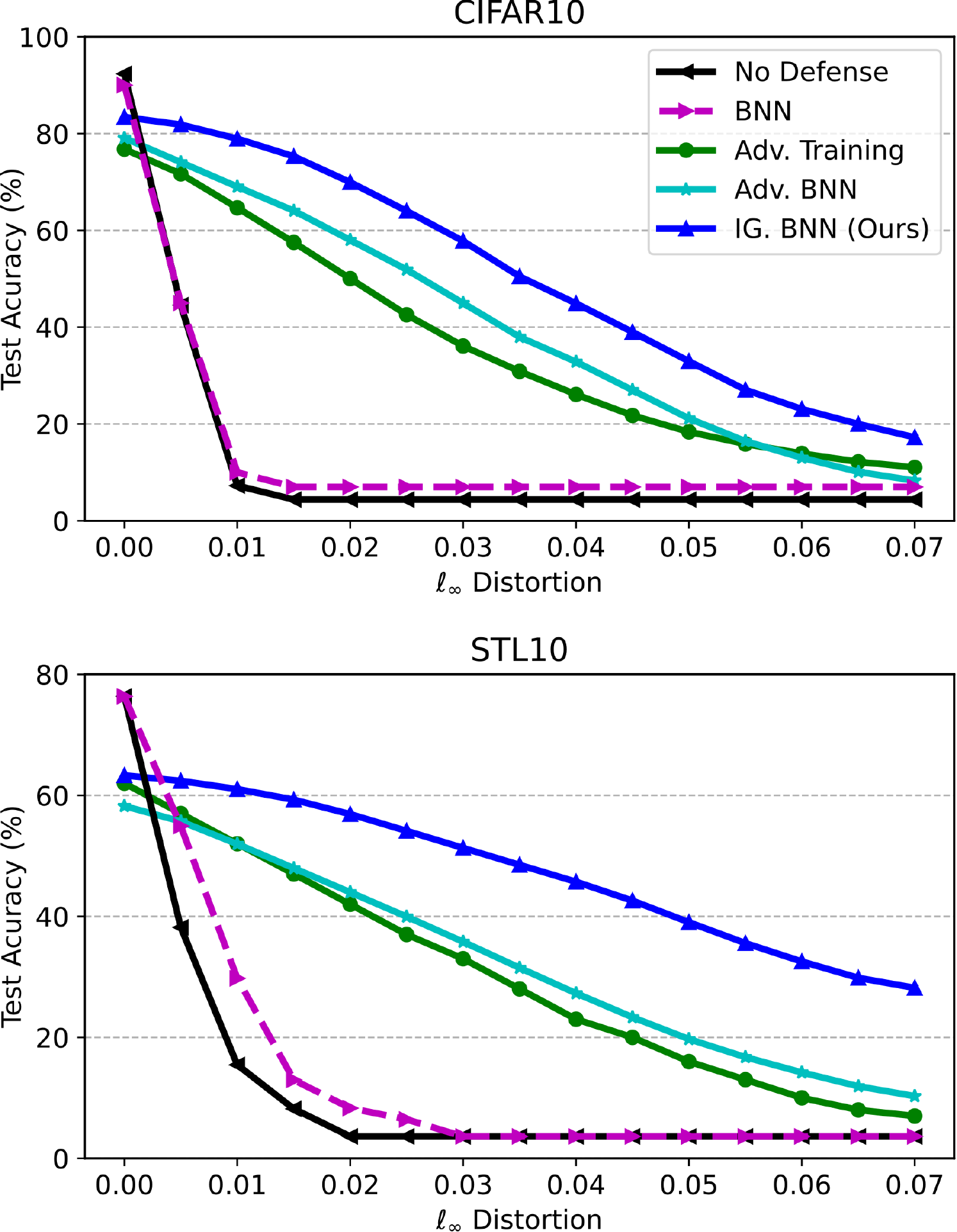}
    \caption{Accuracy under $\ell_\infty$-EoT PGD attack on different datasets. CIFAR-10 is trained on a VGG-16 network, and STL-10 is trained on ModelA--used in Adv-BNN~\cite{advbnn}.}.
    \label{fig:combined}
\end{figure}

\subsection{Robustness Under White-box $l_\infty$ Attacks}

In this experiment, we compare the robustness of our models under the strong white-box $l_\infty$-EoT PGD attack. Following the recent work in~\citep{advbnn}, we set the maximum $l_\infty$ distortion to $\varepsilon_{\max} \in [0:0.07:0.005]$, adjust the PGD attacks for Bayesian methods as mentioned earlier--see Equation~\eqref{eq:a_pgd}---and report the accuracy on the test set (\textit{robustness}). Overall, the results---shown in Figure~\ref{fig:combined}--- illustrate the improved robustness of our method compared with Adv.~BNN~\citep{advbnn}, and the significantly better results compared to  Adv.~Training~\citep{pgd}. We also provide detailed results in Table~\ref{table:apgd_results} where we show a marked increase in testing accuracy (benign) and robustness (against adversarial samples)---notably, IG-BNN achieves up to 17\% at the distortion of 0.035 compared with Adv-BNN and 20\% compared with Adv. Training on STL-10 dataset. These correspond to 13\% on CIFAR-10 and 19\% on STL-10, respectively.   
Although Adv-BNN helped improve robustness, we can see that the learning method is still below what could be achieved. On the other hand, IG-BNN achieved better results on both the testing data (\textit{benign}) and adversarial examples (under increasing attack budgets).

\begin{table}[h!]
\centering
% \small
\caption{\small{Comparing robustness under different levels of EoT PGD attacks (or attack budgets).}}
\label{table:apgd_results}
\begin{adjustbox}{width=\linewidth, center}
\begin{tabular}{ccccccc} 
\dtoprule
\textit{Data} & \textit{Defenses} &  \textit{0} & \textit{0.015} & \textit{0.035} & \textit{0.055} & \textit{0.07}\\
\midrule
\multirow{3}{5em}{CIFAR-10} & Adv. Training & 80.3 & 58.3 & 31.1 & 15.5 & 10.3 \\ 
& Adv-BNN & 79.7 & 64.2 & 37.7 & 16.3 & 8.1 \\ 
& IG-BNN (Ours) & \textbf{83.6} & \textbf{75.5} & \textbf{50.2} & \textbf{26.8} & \textbf{16.9} \\ \midrule 
\multirow{3}{5em}{STL-10} & Adv. Training & 63.2 & 46.7 & 27.4 & 12.8 & 7.0 \\ 
& Adv-BNN  & 59.9 & 47.9 & 31.4 & 16.7 & 9.1 \\
& IG-BNN (Ours) & \textbf{64.3} & \textbf{60.0} & \textbf{48.2} & \textbf{34.9} & \textbf{27.3} \\ %\midrule %double checked

\dbottomrule
\end{tabular}
\end{adjustbox}
\end{table}

\subsection{Ablative Studies}

In this section, we investigate the contribution of each of the formulations in our method. Particularly, we investigate: i)~the contribution of the Bayesian inference method SVGD; and ii)~the contribution of Information Gain (IG). We utilize the same network architecture and training parameters for the \textit{higher resolution}, therefore more challenging, STL-10 dataset with the only difference being the ablative parameter to conduct the experiment. 

\noindent\textbf{Bayesian Inference Methods}.~We evaluate the network trained with the adversarial training using the Bayesian inference method proposed in~\citet{advbnn}, that is Bayes by Backprop (Adv train + BBB), to compare with our  proposed adversarially trained BNN using SVGD (Adv train + SVGD). The results are in Table~\ref{table:ablative_inference}. We can see that employing SVGD with the ability to capture a multi-model posterior contributed to improving the robustness of the Adversarial trained Bayesian Neural Networks.

\begin{table}[h!]
\centering
\small
\caption{\small{Ablative study on assessing the contribution of the Bayesian inference method under different levels of EoT PGD attacks (or attack budgets).}}
\label{table:ablative_inference}
\begin{adjustbox}{width=\linewidth, center}
\begin{tabular}{cccccc} 
\dtoprule
\textit{Defenses} &  \textit{0} & \textit{0.015} & \textit{0.035} & \textit{0.055} & \textit{0.07}\\
\midrule
Adv train + BBB  & 59.9 & 47.9 & 31.4 & 16.7 & 9.1 \\
Adv train + SVGD  & \textbf{63.6} & \textbf{54.2} & \textbf{36.6} & \textbf{24.3} & \textbf{19.4} \\
\dbottomrule
\end{tabular}
\end{adjustbox}
\end{table}

\noindent\textbf{Information Gain}. With the improvements in robustness achieved with the SVGD formulation for adversarial training,  we conduct the ablative study on the network trained with SVGD inference method with and without IG to assess the impact of the IG objective on robustness. 
Notably, the trivial solution for the IG objective is that all parameter particles collapse to a single mode; hence, the IG objective and its effectiveness can be achieved with the inference methods encouraging diversity, such as SVGD.

\begin{table}[h!]
\centering
\small
\caption{\small{Ablative study on assessing the contribution of the Information Gain objective under different levels of EoT PGD attacks (or attack budgets).}}
\label{table:ablative_ig}
\begin{adjustbox}{width=\linewidth, center}
\begin{tabular}{cccccc} 
\dtoprule
\textit{Defenses} &  \textit{0} & \textit{0.015} & \textit{0.035} & \textit{0.055} & \textit{0.07}\\
\midrule
Adv train + SVGD  & 63.6 & 54.2 & 36.6 & 24.3 & 19.4 \\
Adv train + SVGD + IG & \textbf{64.3} & \textbf{60.0} & \textbf{48.2} & \textbf{34.9} & \textbf{27.3} \\ 
\dbottomrule
\end{tabular}
\end{adjustbox}
\end{table}
% \vspace{-2mm}

\begin{figure}[h!]
    \centering
    \includegraphics[width=\linewidth,height=31mm]{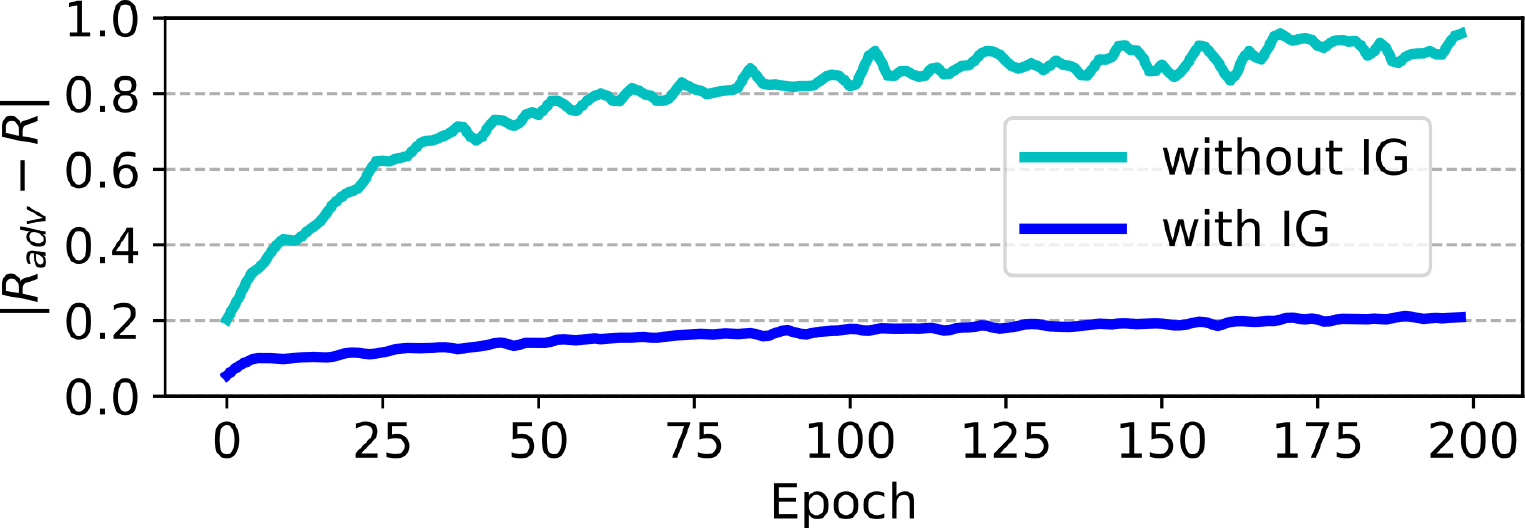}
    \caption{The difference between conventional empirical risk and adversarial risk $\left| R_{\text{adv}}-R \right|$ on the test set is tightened and minimized when training the BNN with Information Gain. Corroborating our proof, the empirical result further explains the improvement in robustness of the IG-BNN networks.}
    \label{fig:Rgap}
\end{figure}

As shown in Table~\ref{table:ablative_ig}, we can see that IG helped improve robustness further, up to 12\%. We also empirically demonstrate the difference in empirical risk and the adversarial risk evaluated on the test set in Figure~\ref{fig:Rgap}. Our empirical results demonstrate the impact of adding IG to tighten the bound and reduce the gap between conventional empirical risk and the adversarial risk; consequently, improving the robustness of the network.

\subsection{Evaluating the Obfuscated Gradient Effect}

One possible failure mode of defense methods discussed in the literature is the obfuscated gradient effect~\citep{athalye2018obfuscated} where seemingly high adversarial accuracy is only superficial and creates false robustness. In this scenario, the network learns to obfuscate the gradients whilst showing seeming robustness by making it harder for the attack to find perturbations. However, an easy and effective way to verify this is to apply a black-box attack on the defense methods. The defense is considered to show an obfuscated gradient effect if the black-box attack is more successful than the white-box attack (\ie the robustness is lower). %Based on the observation and 

Following current practice, in this experiment, we deploy a black-box Square attack~\citep{andriushchenko2020square} on our IG-BNN models. Table~\ref{table:blackbox} shows that our IG-BNN is also highly robust against the black-box attack and, more importantly, the robustness of the black-box attack is significantly higher than the white-box one. Particularly, the robustness against black-box attacks on CIFAR-10 at the distortion of 0.035 is 78.9\% which is a 28\% accuracy improvement compared with its white-box counterpart. On STL-10, at the same distortion, this improvement is 13\%. These results demonstrate that our robustness is not simply the effect of obfuscated gradients.

% \todo{revise the results}
\begin{table}[h!]
\centering
\small
\caption{\small{Blackbox attack to evaluate the obfuscated gradient effect.}}
\label{table:blackbox}
\begin{adjustbox}{width=\columnwidth, center}
\begin{tabular}{ccccccc} 
\dtoprule
\textit{Data} & \textit{Defenses} &  \textit{0} & \textit{0.015} & \textit{0.035} & \textit{0.055} & \textit{0.07}\\
\midrule
\multirow{2}{5em}{CIFAR-10}
& IG-BNN (Ours) & 83.6 & 75.5 & 50.2 & 26.8 & 16.9 \\ 
& Black-box & - & 82.3 & 78.9 & 71.0 & 63.2\\ \midrule
\multirow{2}{5em}{STL-10} 
& IG-BNN (Ours) & 64.3 & 60.0 & 48.2 & 34.9 & 27.3 \\ 
& Black-box & - & 63.8 & 61.3 & 59.3 & 57.6 \\ 
\dbottomrule
\end{tabular}
\end{adjustbox}
\end{table}
% \vspace{-2mm}

\subsection{Transfer Attacks Among Parameter Particles}
To further evaluate the robustness and illustrate the intuition for exploring diverse parameter particles, we conduct experiments on the transferability of the adversarial examples among parameter particles and evaluate the robustness at class-wise levels (i.e. the robustness in each class). 

Specifically, we sample multiple different parameter particles for the experiment. For each parameter particle (\textit{source particles}), we generate corresponding adversarial examples for that parameter particle. And then, using those adversarial examples generated from the source particles, attack and evaluate the robustness of other particles (\textit{target particles}). We visualize the results as heatmaps with robustness as the measure (i.e. the ability to correctly identify the adversarial examples), and show the results in Figure~\ref{fig:transferability}--- We provide comprehensive results in the Appendix~\ref{sec:appd_transfer}. Each row in the matrix shows the robustness of target particles against the adversarial examples generated from the source particles (with the attack budget $\epsilon=0.015$).

As expected, we can observe that the adversarial examples are highly effective on their source particles with 0\% robustness. However, other particles are able to recognize those adversarial examples correctly with high robustness. This further demonstrates the effectiveness of our learning algorithm % is due to our Bayesian learning method
where we encourage the parameter particles to be diverse and additionally bound the difference of empirical risk versus the adversarial risk in terms of the information gain formulation.

\begin{figure}[ht!]
    \centering
    \includegraphics[width=\linewidth]{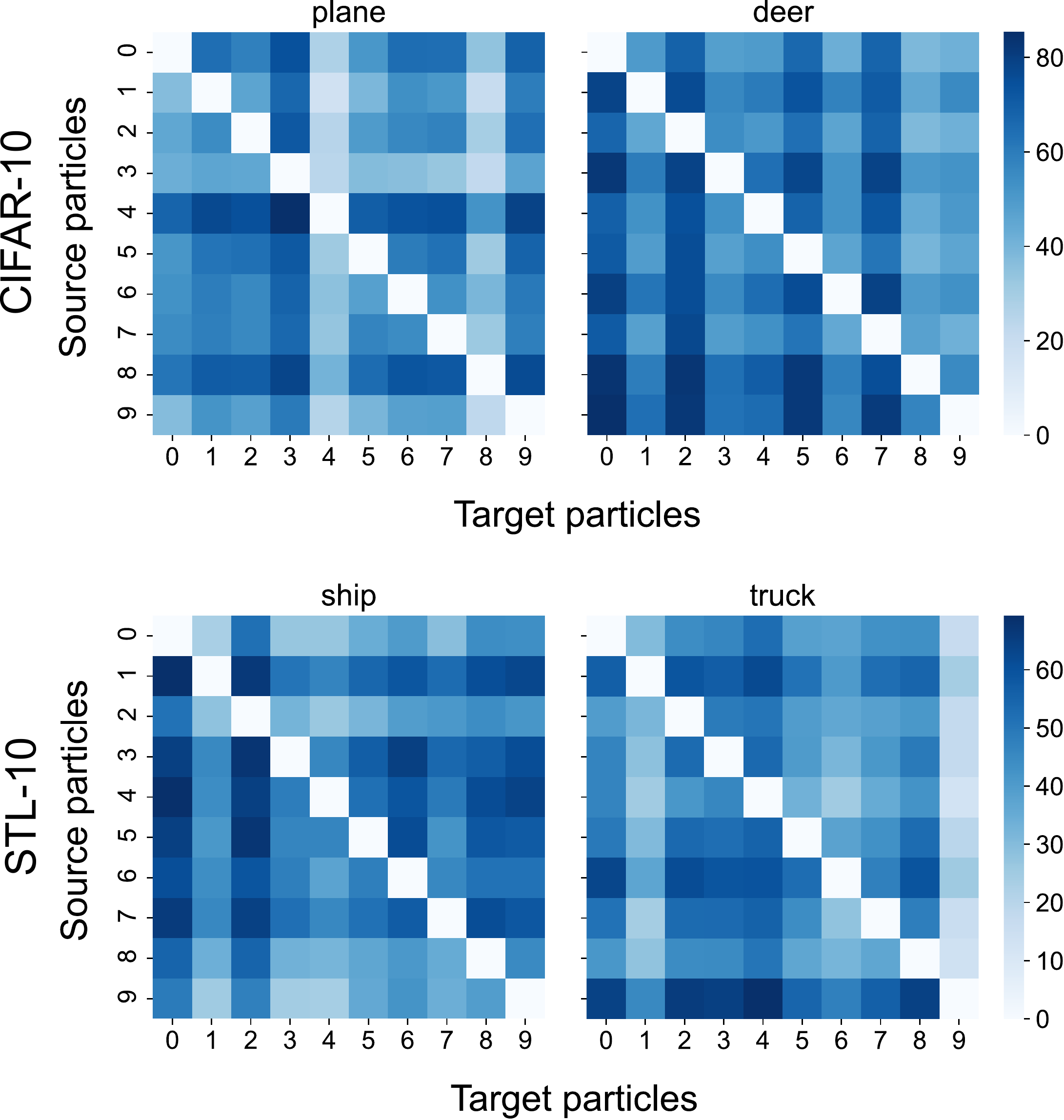}
    \caption{Diversity of parameter particles is demonstrated using the transferability of adversarial examples among particles. We provide comprehensive results in Appendix~\ref{sec:appd_transfer}.}
    \label{fig:transferability}
\end{figure}

% \vspace{-2mm}
\section{Conclusion}

In this study, we presented a novel method to learn a robust BNN against adversarial attacks. We demonstrate that, although an adversarially trained BNN improved robustness, the improvement is slight compared with the traditional adversarial training when using the EoT PGD attack tailored for BNNs. Our proposed IG-BNN learning method employing SVGD to encourage diverse parameter particles together with the formulated information gain objective under the Bayesian context provably bounds the difference of empirical risk versus adversarial risk to yield improved robustness.
The empirical experiments demonstrate that learning a Bayesian neural network using our method tightens the gap between the empirical risk and the empirical adversarial risk; this, consequently leads to better robustness compared with previous adversarially trained Bayesian defense methods.

\bibliography{ref}
\bibliographystyle{icml2022}
%%%%%%%%%%%%%%%%%%%%%%%%%%%%%%%%%%%%%%%%%%%%%%%%%%%%%%%%%%%%

\appendix
\onecolumn
\newpage

\section{Definition of Information Gain}
\label{sec:appd_proofdefinition}
We first define our predictive distribution as:
$$p(y|\bx,\D)=\int p(y|\bx,\btheta)p(\btheta|\D)d\btheta\,.$$

% \tocheck{change the sign of Information Gain}

Following the definition of information gain, we have:
\begin{align*}
\mathbb{E}[\text{IG}(\bx,y;\bTheta)] &=\sum_{y}p(y|\bx,\D))\int\frac{p(y|\bx,\boldsymbol{\theta})p(\D|\btheta)p(\btheta)}{p(\D)p(y|\bx,\D)}\log\left(\frac{p(y|\bx,\btheta)}{p(y|\bx,\D)}\right)d\btheta \\
	&=	\frac{1}{p(\D)}\sum\int p(y|\bx,\btheta)p(\D|\btheta)p(\btheta)\log\left(\frac{p(y|\bx,\btheta)}{p(y|\bx,\D)}\right)d\btheta \\
	&=	\frac{1}{p(\D)}\sum\int p(y|\bx,\btheta)p(\btheta|\D)\log\left(\frac{p(y|\bx,\btheta)}{p(y|\bx,\D)}\right)d\btheta \\
	&=	\frac{1}{p(\D)}\sum\int p(y|\bx,\btheta)p(\btheta|\D)\left[\log(p(y|\bx,\btheta))-\log(p(y|\bx,\D))\right]d\btheta \\
	&=	\frac{1}{p(\D)}\sum\left[\int p(y|\bx,\btheta)p(\btheta|\D)\log(p(y|\bx,\btheta))d\btheta-\int  p(y|\bx,\btheta)p(\btheta|\D)\log(p(y|\bx,\D))d\btheta\right] \\
	&=	\frac{1}{p(\D)}\int p(\btheta|\D)\sum p(y|\bx,\btheta)\log(p(y|\bx,\btheta))d\btheta-\sum\int p(y|\bx,\btheta)p(\btheta|\D)\log(p(y|\bx,\D))d\btheta \\
	&=	\frac{1}{p(\D)}\left(\mathbb{H}[\mathbb{E}_{\btheta}[y|\bx,\D]]-\mathbb{E}_{\btheta}[\mathbb{H}[y|\bx,\D]]\right)\\
	&\propto \bigg(\mathbb{H}[\mathbb{E}_{\btheta}[y|\bx,\D]]-\mathbb{E}_{\btheta}[\mathbb{H}[y|\bx,\D]]\bigg)
\end{align*}
where for the last line we assume $p(\D)\approx p(\Da)$ as constant values. Since we are considering adversarial instances to be obtained from the observational one, this is a very mild assumption and is completely in alignment with  current research.

\section{Proof of the Objective}
\label{sec:appd_proof}
We have
\begin{align*}\left|R_{\text{adv}}-R\right| & =\left|\mathbb{E}_{(\bx,y)\sim\D}\Bigg[\mathbb{E}_{\btheta}\Bigg[\underset{}{\sup}\,\,\mathbb{E}_{y_{1}\sim p(y|\bxa)}\left[\II\left(y_{1}\neq y\right)\right]-\mathbb{E}_{y_{2}\sim p(y|\bx)}\left[\II\left(y_{2}\neq y\right)\right]\Bigg]\Bigg]\right|\,,\\
 & =\left|\mathbb{E}_{(\bx,y)\sim\D}\Bigg[\mathbb{E}_{\btheta}\Bigg[\sup\,\,\mathbb{E}_{y_{1}\sim p(y|\bxa),y_{2}\sim p(y|\bx)}\left[\II\left(y_{1}\neq y\right)-\II\left(y_{2}\neq y\right)\right]\Bigg]\Bigg]\right|\,,\\
 & \quad\left.\leq\mathbb{E}_{(\bx,y)\sim\D}\Bigg[\mathbb{E}_{\btheta}\Bigg[\sup\,\,\mathbb{E}_{y_{1}\sim p(y|\bxa),y_{2}\sim p(y|\bx)}\left[|\II\left(y_{1}\neq y\right)-\II\left(y_{2}\neq y\right)|\right]\Bigg]\right]\,,\\
 & \quad\leq\mathbb{E}_{(\bx,y)\sim\D}\Bigg[\mathbb{E}_{\btheta}\Bigg[\sup\,\,\mathbb{E}_{y_{1}\sim p(y|\bxa),y_{2}\sim p(y|\bx)}\left[\II\left(y_{1}\neq y_{2}\right)\right]\Bigg]\Bigg]\,.
 \end{align*}
 where we can upper bound the expected misclassification to have:
 \begin{align*}
 &\mathbb{E}_{(\bx,y)\sim\D}\Bigg[\mathbb{E}_{\btheta}\bigg[1-\sum_{c=1}^{K}p(y=c\mid\bx,\btheta)p(y=c\mid\bxa,\btheta)\bigg]\Bigg]\,.
 \end{align*}
 Subsequently, we use Jensen's inequality and the fact that $\bx=\exp(\log(\bx))$ to have:
 \begin{align*}
 &\mathbb{E}_{(\bx,y)\sim\D}\Bigg[\mathbb{E}_{\btheta}\bigg[1-\exp(\underbrace{\log(\sum_{c=1}^{K}p(y=c\mid\bx,\btheta)p(y=c\mid\bxa,\btheta))}_{\geq \sum_{c}^{K}p(y=c\mid\bx,\btheta)\log(p(y=c\mid\bxa,\btheta)})\bigg]\Bigg]\,.
 \end{align*}
 
 For a monotonically decreasing function, we know for $x\geq y$, $f(x)\leq f(y)$.
Using Jensen's inequality we have
${\log(\sum_{c=1}^{K}p(y=c\mid\bx,\btheta)p(y=c\mid\bxa,\btheta))}{\geq \sum_{c}^{K}p(y=c\mid\bx,\btheta)\log(p(y=c\mid\bxa,\btheta)}$. Since $1-\exp(z)$ is monotonically decreasing, 
we have:
 \begin{align*}
 &\mathbb{E}_{(\bx,y)\sim\D}\Bigg[\mathbb{E}_{\btheta}\bigg[1-\exp({\log(\sum_{c=1}^{K}p(y=c\mid\bx,\btheta)p(y=c\mid\bxa,\btheta))})\bigg]\Bigg]\\
 & \leq \mathbb{E}_{(\bx,y)\sim\D}\Bigg[\mathbb{E}_{\btheta}\bigg[1-\exp\big(\sum_{c}^{K}p(y=c\mid\bx,\btheta)\log(p(y=c\mid\bxa,\btheta))\big)\bigg]\Bigg]\\
 & \quad\quad= 1-\mathbb{E}_{(\bx,y)\sim\D}\Bigg[\mathbb{E}_{\btheta}\bigg[\exp\big(\sum_{c}^{K}p(y=c\mid\bx,\btheta)\log(p(y=c\mid\bxa,\btheta))\big)\bigg]\Bigg]\,.\\
 & \qquad\quad
\end{align*}

Thus we have the following bound:
\begin{align}
\label{eq:initial_bound}
\left|R_{\text{adv}}-R\right|\leq1-\mathbb{E}_{(\bx,y)\sim\D}\Bigg[\exp\Bigg(\mathbb{E}_{\btheta}\bigg[\underbrace{\sum_{c}^{K}p(y=c\mid\bx,\btheta)\log(p(y=c\mid\bxa,\btheta))}_{r_{\btheta}(\bx,\bxa,y)}\bigg]\Bigg)\Bigg]\,.
\end{align}
This result demonstrates that the difference between the risks is bounded by the negative cross-entropy of the predictions. While informative, this bound expresses the  relation between the predictions only and not how the model performs on each set (\ie given dataset versus its corresponding adversarial).

From the definition of KL-divergence, we know
$$r_{\btheta}(\bx,\bxa,y) = -\mathbb{H}(p(y=c\mid\bx,\btheta), p(y=c\mid\bxa,\btheta))= -{\mathrm{KL}}(p(y=c\mid\bx,\btheta) \| p(y=c\mid\bxa,\btheta) -\mathbb{H}(p(y=c\mid\bx,\btheta))$$

We can add and subtract $\mathbb{H}[\mathbb{E}_{\btheta}[p(y=c\mid\bx,\btheta)]]$ and $\mathbb{E}_{\btheta}[\text{IG}(\bxa,y)]$ to have
{\vspace{-0mm}\small\begin{align*}
\vspace{-0mm}
     \mathbb{E}_{\btheta}[r_{\btheta}(\bx,\bxa,y)] &= -\mathbb{E}_{\btheta}[{\mathrm{KL}}(p(y=c\mid\bx,\btheta) \| p(y=c\mid\bxa,\btheta))]- \mathbb{H}[\mathbb{E}_{\btheta}[p(y=c\mid\bx,\btheta)]]+\mathbb{E}_{\btheta}[\text{IG}(\bxa,y)] \vspace{-3mm} \\
    &\qquad\qquad +\underbrace{(\mathbb{H}[\mathbb{E}_{\btheta}[p(y=c\mid\bx,\btheta)]]-\mathbb{E}_{\btheta}[\mathbb{H}[p(y=c\mid\bx,\btheta)]])}_{\mathbb{E}_{\btheta}[\text{IG}(\bx,y)]}-\mathbb{E}_{\btheta}[\text{IG}(\bxa,y)] \\
    &\vspace{-10mm}=-\mathbb{E}_{\btheta}[{\mathrm{KL}}(p(y=c\mid\bx,\btheta) \| p(y=c\mid\bxa,\btheta))]-\underbrace{(\mathbb{E}_{\btheta}[\text{IG}(\bxa,y)]-\mathbb{E}_{\btheta}[\text{IG}(\bx,y)])}_{A}\\
    &\vspace{-10mm}\quad+\underbrace{\mathbb{E}_{\btheta}[\text{IG}(\bxa,y)] - \mathbb{H}[\mathbb{E}_{\btheta}[p(y=c\mid\bx,\btheta)]]}_{B}=-\mathbb{E}_{\btheta}[{\mathrm{KL}}(p(y=c\mid\bx,\btheta) \| p(y=c\mid\bxa,\btheta))]- A + B.
\end{align*}}%

We consider two cases: 

\begin{itemize}[label={}]
    \item \textbf{i)} $A=0$, then
$\mathbb{E}_{\btheta}[r_{\btheta}(\bx,\bxa,y)]=-\mathbb{E}_{\btheta}[{\mathrm{KL}}(p(y=c\mid\bx,\btheta) \| p(y=c\mid\bxa,\btheta))] -\mathbb{E}_{\btheta}[\mathbb{H}(p(y=c\mid\bx,\btheta))] \leq -\mathbb{E}_{\btheta}[{\mathrm{KL}}(p(y=c\mid\bx,\btheta) \| p(y=c\mid\bxa,\btheta))]$, because $\mathbb{E}_{\btheta}[\mathbb{H}(p(y=c\mid\bx,\btheta))]\geq0$. Therefore, in this case, 
   $-\mathbb{E}_{\btheta}[{\mathrm{KL}}(p(y=c\mid\bx,\btheta) \| p(y=c\mid\bxa,\btheta))]$ is an upper bound on $\mathbb{E}_{\btheta}[r_{\btheta}(\bx,\bxa,y)]$.
   
    \item \textbf{ii)} $A\neq 0$, then we have $-A + B=A(-1+B/A)$. We know $A\leq|A|$ for any value, then $A(-1+B/A)\leq |A|(-1+B/A)$. Setting $(-1+B/A)=-\lambda$, we have $\lambda=(1-B/A)$. In practice, we tune $\lambda$ as detailed in the paper. As such, we have, $-A + B \leq -\lambda |A|$. 
\end{itemize}

Thus, putting case (i) and (ii) together, we have:
\begin{equation}
\mathbb{E}_{\btheta}[r_{\btheta}(\bx,\bxa,y)] \leq -\mathbb{E}_{\btheta}[{\mathrm{KL}}(p(y=c\mid\bx,\btheta) \| p(y=c\mid\bxa,\btheta))] - \lambda |\mathbb{E}_{\btheta}[\text{IG}(\bx,y;\bTheta)]-\mathbb{E}_{\btheta}[\text{IG}(\bxa,y;\bTheta)]|
\end{equation}
and since $1-\exp(\cdot)$ is monotonically decreasing, we are able to achieve a tighter bound for Eq.~\eqref{eq:initial_bound} with: 

{\small\begin{equation}
\label{eq:final_bound}
\left|R_{\text{adv}}-R\right| \leq 1-\mathbb{E}_{(\bx,y)\sim\D}\Bigg[\exp\bigg({-\big(\mathbb{E}_{\btheta}[\text{KL}(p(y=c\mid\bx,\btheta)\|p(y=c\mid\bxa,\btheta))]+\lambda|\mathbb{E}_{\btheta}[\text{IG}(\bx,y;\bTheta)]-\mathbb{E}_{\btheta}[\text{IG}(\bxa,y;\bTheta)]}|\big)\bigg)\Bigg]\,.
\end{equation}}
Then the difference between the empirical risk and the adversarial risk is minimized when the upper bound (the right-hand expression of the Eq.~\eqref{eq:final_bound}) is minimized. Hence, the main learning objectives are to:
\begin{enumerate}
    \item Minimize $\mathbb{E}_{\btheta}[\text{KL}(p(y=c\mid\bx,\btheta)\|p(y=c\mid\bxa,\btheta))]$: This corresponds to matching the prediction from the adversarial data to that of the observations. Since $(\bx,y)$ is given in training, for minimizing this KL-divergence we simply convert the minimization of the KL term to minimization of the cross-entropy loss of the adversarial examples instead.
    
    \item Minimize $|\mathbb{E}_{\btheta}[\text{IG}(\bx,y;\bTheta)]-\mathbb{E}_{\btheta}[\text{IG}(\bxa,y;\bTheta)]|$: In addition to individual predictions, the information gained from each instance has to have a similar impact on the network in terms of how it changes the parameters.
\end{enumerate}
Notably, since we know $1-\exp(-z)\leq z$, to avoid computational instabilities and gradient saturation, we consider minimizing the upper bound without the exponential function in our implementation.

% \newpage

\section{Hyper-Parameters}
\label{sec:appd_hyper}

These are hyper parameters used in our experiments. For a fair comparison with previous works, all of the training, testing parameters and attack budgets are identical to those in~\citet{advbnn}.

\begin{table}[h!]
\centering
\small
\caption{\small{Hyper-parameters setting in our experiments}}
\label{table:hyperparameters}
\begin{adjustbox}{width=0.55\linewidth, center}
\begin{tabular}{ccc} 
\dtoprule
\textit{Name} & \textit{Value} &  \textit{Notes} \\
\midrule
$T'$ & 20 & \#PGD iterations in attack at test time \\ \midrule
$T$ & 10 & \#PGD iterations in adversarial training \\ \midrule
$\varepsilon_{\max}$ & 8/255 & Max $l_\infty$-norm in adversarial training \\ \midrule
$\alpha$ & 2/255 & Step size for each PGD iteration \\ \midrule
$\gamma$ & 0.01 & Weight to control the repulsive force \\ \midrule
$\lambda$ & CIFAR-10: 5, STL-10: 20 & Weight to control IG objective \\ \midrule
$n$ & 10 & \makecell{\#Parameter particles \\ \#Forward passes when doing ensemble inference \\ \# Expectation over Transformation} \\ \dbottomrule

\end{tabular}
\end{adjustbox}
\end{table}

\section{Experiment with Increasing Number of EoT-PGD Steps}

\begin{figure}[h!]
    \centering
    \includegraphics[width=0.5\linewidth]{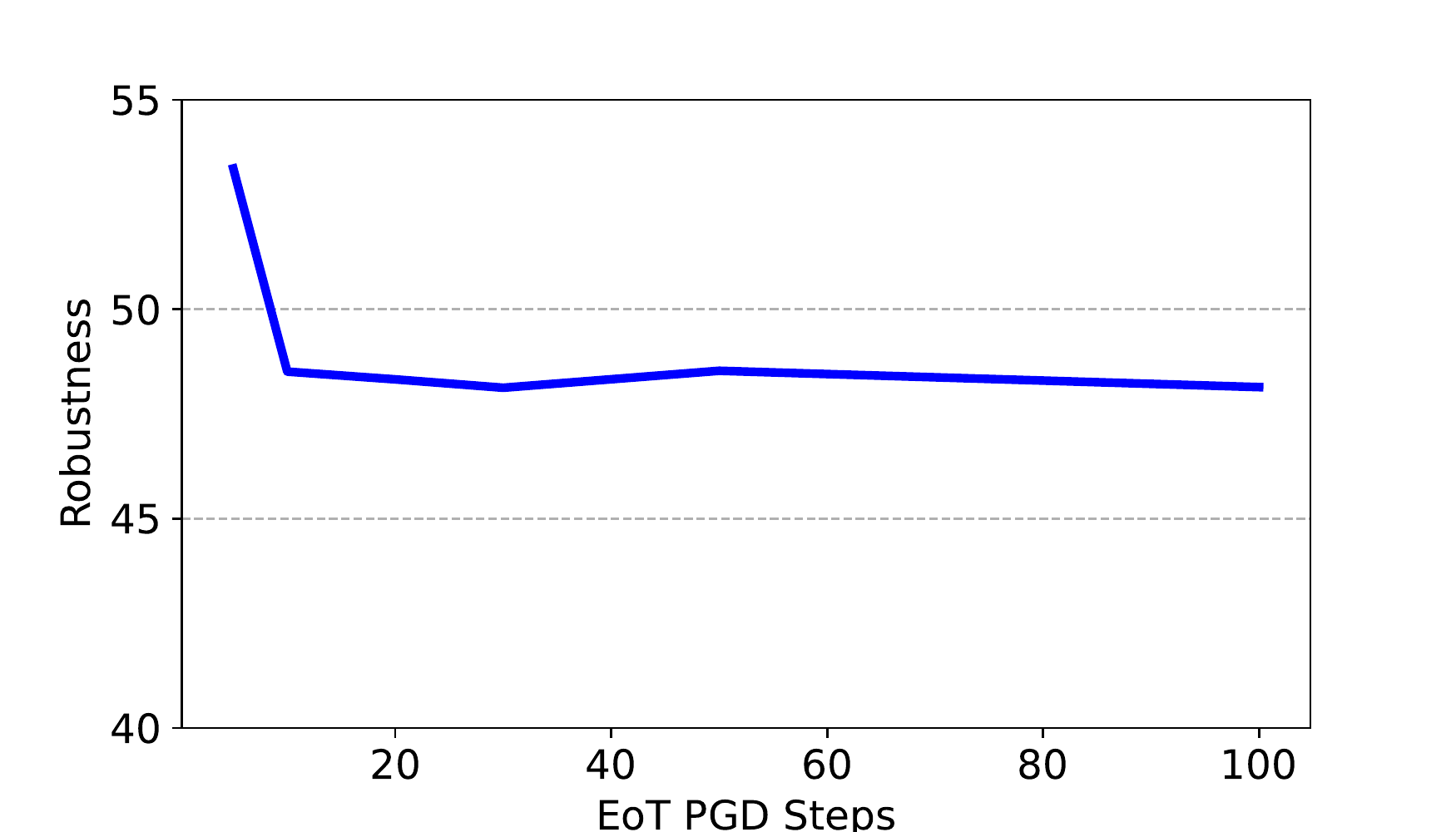}
    \caption{Robustness versus various numbers of EoT-PGD steps. EoT-PGD reaches its full strength after 20 steps. Further increasing PGD steps did not significantly improve the attack.}
    \label{fig:PGDsteps}
\end{figure}

Following standard practice and due to the cost of running increasing numbers of EoT-PGD steps, the results in the main paper use 20 steps. In this section, we conduct experiments with an increasing number of EoT-PGD steps to demonstrate  that the robustness we evaluated in the main paper is on a so-called full-strength EoT-PGD. As shown in Figure~\ref{fig:PGDsteps}, the robustness is significantly decreased in the first 20 steps. However, after that, robustness is maintained, \ie the EoT-PGD attack has converged and reached its full strength. 
\label{sec:appd_PGDsteps}

\FloatBarrier
\section{Transferability to Other Attacks}
In this section, in order to extend the scope of the method and to show that our method is generic and applicable to other adversarial attacks, we conduct experiments to evaluate the robustness of networks trained on EoT-PGD $\ell_\infty$ against different attacks such as FGSM or $\ell_2$-attack. Results in Table~\ref{table:other_adversaries} show that our method's robustness is transferable to other attacks. The reason is that we utilized PGD in our method, and PGD is regarded as a ``universal" adversary among first-order approaches, i.e. if a network is robust against PGD adversaries, it will be robust against a wide range of other attacks~\citep{pgd}.

\begin{table}[h]
\centering
\caption{\small{Transferability. PGD $\ell_\infty$ trained IG-BNN robustness  against different adversaries under different attack budgets.}}
\label{table:other_adversaries}
{\small\begin{adjustbox}{width=.5\linewidth, center}
\begin{threeparttable}
\begin{tabular}{cccccc} 
\dtoprule
\textit{Attacks on} \texttt{CIFAR-10} &  \textit{0} & \textit{0.015} & \textit{0.035} & \textit{0.055} & \textit{0.07}\\
\midrule
PGD $\ell_\infty$  & 83.6 & 75.5 & 50.2 & 26.8 & 16.9 \\
FGSM  & - & 76.1  & 55.7 & 38.4 & 28.9 \\
PGD $\ell_2$  & - & 83.5  & 83.4 & 83.2 & 83.1 \\
\dbottomrule
\end{tabular}
\end{threeparttable}
\end{adjustbox}}%
% \vspace{-5mm}
\end{table}

\section{Validating Our Conjecture}
\begin{figure}[h]
    \centering
    \includegraphics[width=.8\linewidth]{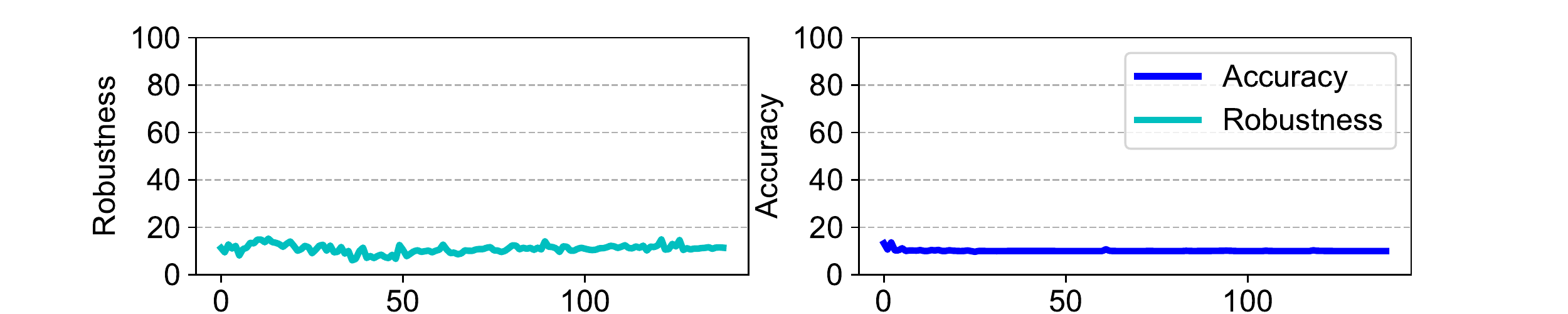}
    \caption{Accuracy and Robustness of the BNN network trained on STL-10 dataset where we enforce the model to be `inconsistent' under clean settings and adversarial settings.}
    \label{fig:conjecture}
\end{figure}

Our method is built upon the conjecture: \textit{a robust neural network quantifies the information gained from observation the same as its adversarial counterpart}. In this section, we further support this conjecture by conducting an evaluation where we assess the opposite conjecture. We make the BNN model `inconsistent' under clean settings and adversarial settings. More specifically,  instead of minimizing the Information Gain objective, we maximize to enforce the inconsistency. Figure~\ref{fig:conjecture} shows that this inconsistency leads to the deterioration of the network's performance. This experiment empirically validates our conjecture.   

\newpage

\section{Details of Transfer Attacks of Adversarial Examples Among Parameter Particles}
\label{sec:appd_transfer}

\begin{figure}[h!]
    \centering
    \includegraphics[width=\textwidth]{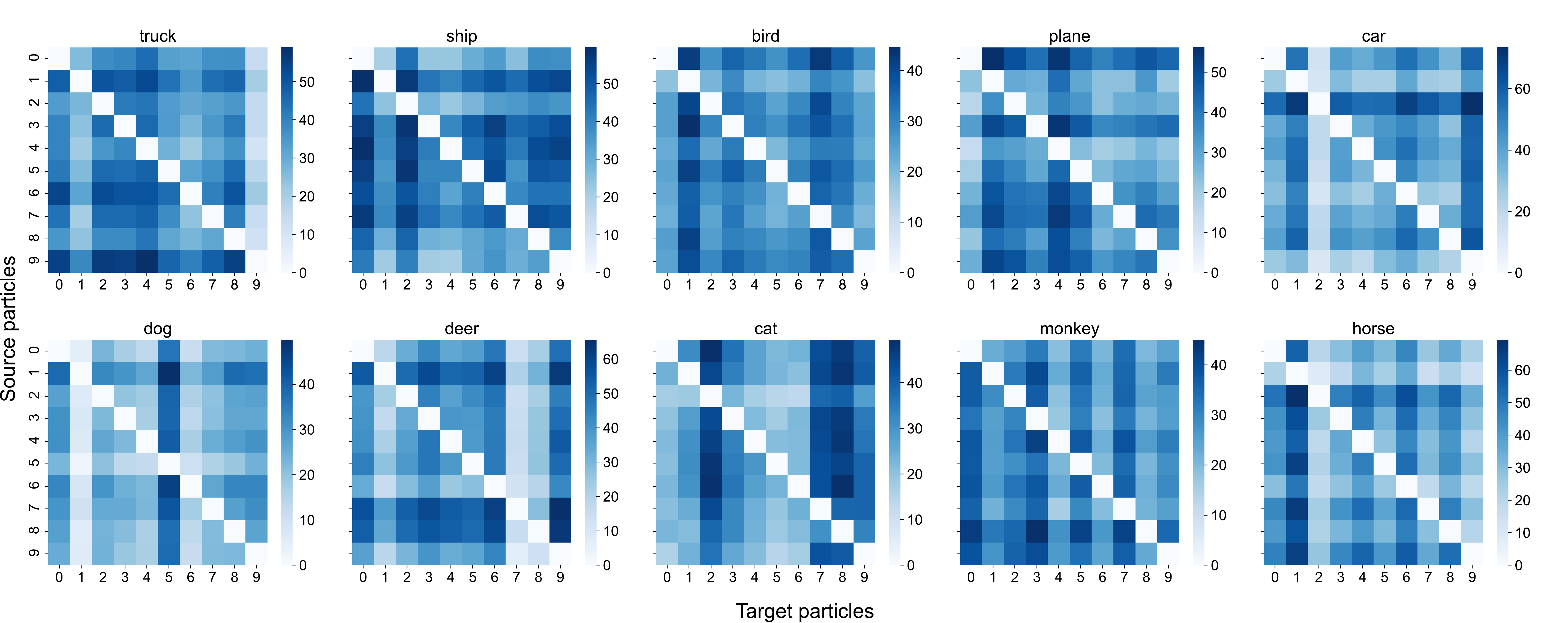}
    \caption{Transferability of adversarial examples among different particles on STL-10}
    \label{fig:trans_stl}
\end{figure}
\begin{figure}[h!]
    \centering
    \includegraphics[width=\textwidth]{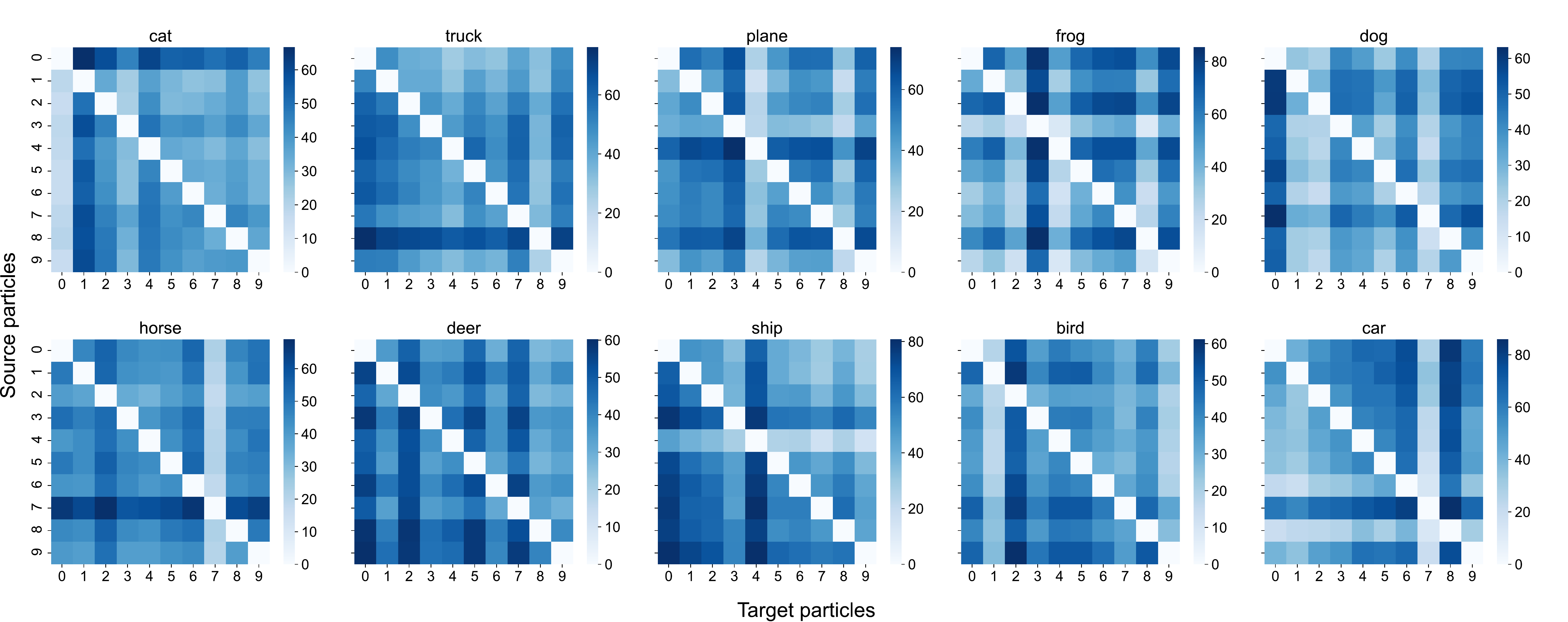}
    \caption{Transferability of adversarial examples among different particles on CIFAR-10}
    \label{fig:trans_cifar}
\end{figure}

\end{document}